\documentclass{article}
\usepackage{spconf,amsmath,graphicx,hyperref}
\usepackage{amssymb}
\usepackage{cite}

\title{Forward-Only Learning for Shared-Kernel Wavelet Neural Networks without Backpropagation}
\label{key}\name{
	Yongfei Guo$^{1,2}$,
	Xuan Sun$^{1,2}$,
	Ang Li$^{1,3}$,
	Zhao Zhang$^{1,3}$,
	Yuanhao Gong$^{1,3}$
	\thanks{This work was supported in part by the National Natural Science Foundation of China under Grant 12471502, the CAS Hundred Talents Program, and the Science and Technology Development Plan Project of Jilin Province, China under Grant 20260204053YY.}
}

\address{
	\textit{$^{1}$Changchun Institute of Optics, Fine Mechanics and Physics, Chinese Academy of Sciences}\\
	\textit{$^{2}$University of Chinese Academy of Sciences, Beijing, China}\\
	\textit{$^{3}$Chinese Academy of Sciences, Beijing, China}
}
\begin{document}
%\ninept
%
\maketitle
\begin{abstract}
	%Backpropagation remains the dominant training paradigm, using backward gradients and local updates that can suffer from vanishing or exploding gradients in deep networks. Although the Forward-Forward algorithm avoids conventional backpropagation, it remains mainly local or layer-wise, motivating global whole-network learning using forward evaluations only. For shared-kernel wavelet neural networks, we therefore propose a CMA-ES-based global forward-only learning framework. We further introduce structure-aware parameter decomposition and channel-specific credit assignment, factorizing the search into three aligned 59-dimensional subspaces. On BSDS500, the proposed method reduces test MSE from 4.1783 to 4.0130 and improves PSNR from 42.621 to 42.787~dB over Adam/BP without backward propagation. Ablations show that structural decomposition and aligned credit assignment are more important than increasing population size, demonstrating the feasibility of whole-network forward-only learning for compact reconstruction networks. These results suggest a potential route toward forward-only optimization of compact reconstruction models for resource-constrained imaging and edge or near-sensor systems.
	
	Backpropagation remains the dominant training paradigm, using backward gradients and local updates that can suffer from vanishing or exploding gradients in deep networks. Although the Forward-Forward algorithm avoids conventional backpropagation, it remains mainly local or layer-wise, motivating global whole-network learning using forward evaluations only. For shared-kernel wavelet neural networks, we therefore propose a CMA-ES-based global forward-only learning framework. We further introduce structure-aware parameter decomposition and channel-specific credit assignment, factorizing the search into three subspaces. On BSDS500, our method reduces test MSE from 4.1783 to 4.0130 and improves PSNR from 42.621 to 42.787~dB over Adam/BP without backward propagation. Beyond these gains, training requires only 6 training-set coverages, compared with 1500 epochs for BP. Ablations show that structural decomposition and aligned credit assignment are more important than increasing population size, demonstrating the feasibility of whole-network forward-only learning for compact reconstruction networks. These results suggest a potential route toward forward-only optimization of compact reconstruction models for resource-constrained imaging and edge or near-sensor systems.
\end{abstract}

\begin{keywords}
	Forward-only learning, CMA-ES, shared-kernel networks, image reconstruction
\end{keywords}

\section{Introduction}
\label{sec:introduction}

Image reconstruction is a fundamental problem in computational imaging and low-level vision, where neural methods recover high-quality images from degraded or transformed measurements~\cite{zamir2022restormer,chen2022nafnet,potlapalli2023promptir}. Practical systems increasingly require not only reconstruction accuracy but also low memory and computational costs. Multiscale and wavelet-inspired architectures are well suited to this goal because they capture image structures across resolutions while introducing structured analysis--synthesis operations~\cite{wang2022uformer,huang2022winnet,deng2023pirnet}.

\begin{figure}[t]
	\centering
	\includegraphics[width=\columnwidth]{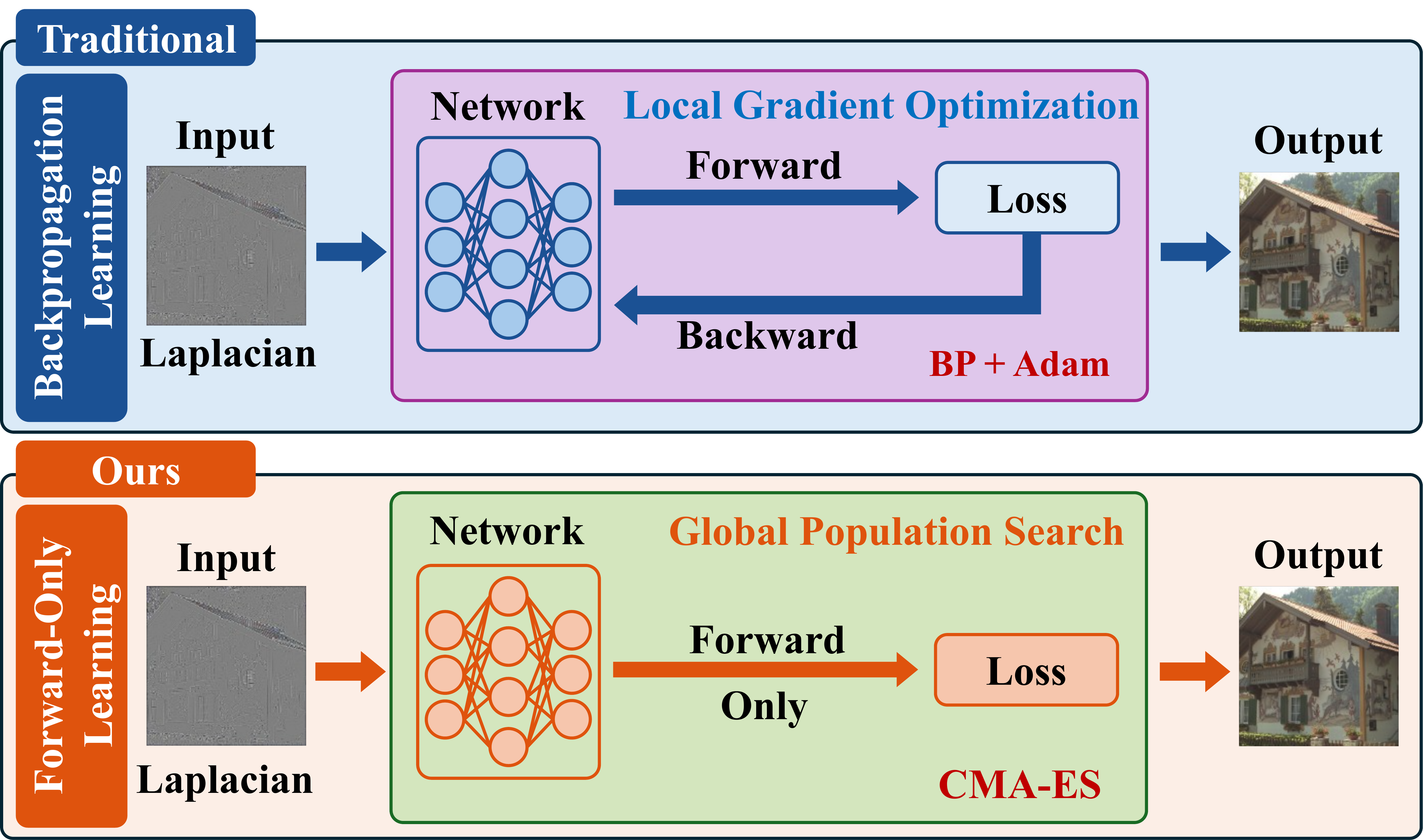}
	\caption{\textbf{Local} gradient-based optimization versus \textbf{global} forward-only population search. BP uses backward gradients for local updates. In contrast, our method evaluates complete parameter candidates through forward losses and performs population-based global search with CMA-ES.}
	\label{fig:learning_paradigm}
\end{figure}

Shared-kernel wavelet networks combine multiscale analysis--synthesis with cross-scale parameter sharing, yielding a compact trainable parameterization while retaining reconstruction capability~\cite{gong2026sharedkernel}. Beyond model efficiency, such a low-dimensional parameter space can make optimization strategies impractical for large networks tractable.

Backpropagation remains the dominant neural-network training paradigm, using local first-order gradient information for iterative parameter updates~\cite{radhakrishnan2024backpropfree,lorberbom2024layer}. Recent backpropagation-free methods, including the Forward-Forward algorithm, avoid conventional backward propagation but remain predominantly based on layer-local objectives~\cite{hinton2022forwardforward,ren2023forwardgradient,lorberbom2024layer}. In contrast, evolutionary and derivative-free methods evaluate populations of complete candidate solutions, enabling global exploration without gradients~\cite{li2023evolutionaryml,hansen2016cma}. Here, global refers to population-based search over the complete network parameterization rather than guaranteed convergence to a global optimum. Although such whole-network search is generally impractical for large neural networks, the compact parameterization of shared-kernel wavelet networks makes it tractable.

In this work, we propose a global forward-only learning framework for shared-kernel wavelet neural networks. As illustrated in Fig.~\ref{fig:learning_paradigm}, conventional BP relies on backward gradient propagation and local gradient-based updates, whereas our approach evaluates complete network candidates through forward reconstruction losses and adapts a population distribution using CMA-ES~\cite{hansen2016cma}. Compactness alone, however, does not make naive joint search effective. We therefore introduce structure-aware parameter decomposition and channel-specific credit assignment to align the search with the intrinsic network structure. The resulting framework optimizes the complete network through forward evaluations only, without automatic differentiation or backward propagation.

The main contributions are summarized as follows:
\begin{itemize}
	\item We introduce global forward-only learning for compact shared-kernel wavelet networks, replacing backward gradients with population-based search.
	\item We develop structure-aware optimization with parameter decomposition and channel-specific credit assignment for effective whole-network learning.
	\item We validate the framework through BP comparisons, structural ablations, and multi-seed evaluations, demonstrating stable reconstruction without backpropagation.
\end{itemize}

\section{Related Work}
\label{sec:related_work}

We review two lines of work relevant to our framework: compact multiscale reconstruction networks for parameter-efficient architectures and backpropagation-free learning as an alternative to conventional gradient-based training.
\subsection{Compact Multiscale Reconstruction Networks}
\label{subsec:compact_multiscale}

Multiscale reconstruction networks exploit hierarchical representations to recover image structures across resolutions. Uformer, Restormer, and NAFNet demonstrate effective multiscale or efficient restoration designs~\cite{wang2022uformer,zamir2022restormer,chen2022nafnet}, while wavelet-inspired models such as WINNet and PIRNet explicitly incorporate multiresolution signal-processing structures~\cite{huang2022winnet,deng2023pirnet}. Recent lightweight models further reduce complexity through parameter-efficient transformations~\cite{wu2024compacter}. Shared-kernel wavelet networks extend this direction by reusing a small set of learnable kernels across scales~\cite{gong2026sharedkernel}. Whereas existing studies primarily exploit compactness for architectural efficiency, we investigate whether extreme parameter sharing can also enable a different learning regime.

\subsection{Backpropagation-Free Neural Learning}
\label{subsec:backprop_free_learning}

Backpropagation-free methods replace end-to-end gradient propagation with alternative learning signals. Forward-Forward and PEPITA use layer-wise objectives or error-driven forward modulation~\cite{hinton2022forwardforward,dellaferrera2022pepita}, with subsequent extensions remaining predominantly local or layer-wise~\cite{ren2023forwardgradient,lorberbom2024layer,papachristodoulou2024cwcomp,srinivasan2024forward,chen2025selfcontrastive,sun2025deeperforward}. MeZO performs zeroth-order optimization using forward evaluations~\cite{malladi2023mezo}, while evolutionary methods evaluate populations of complete candidates~\cite{li2023evolutionaryml}. CMA-ES adapts the search mean, covariance, and step size for continuous black-box optimization~\cite{hansen2016cma}, but whole-network search becomes difficult in high dimensions. We instead exploit extreme network compactness to make such search tractable and structure-aware.

\section{Global Forward-Only Learning}
\label{sec:forward_only_learning}

Our framework combines a compact shared-kernel wavelet network with CMA-ES-based forward-only search. We first define the shared-kernel reconstruction model, then introduce population-based search and the structure-aware decomposition with channel-specific credit assignment.

\subsection{Shared-Kernel Wavelet Network}
\label{subsec:shared_kernel_network}

\begin{figure}[t]
	\centering
	\includegraphics[width=\columnwidth]{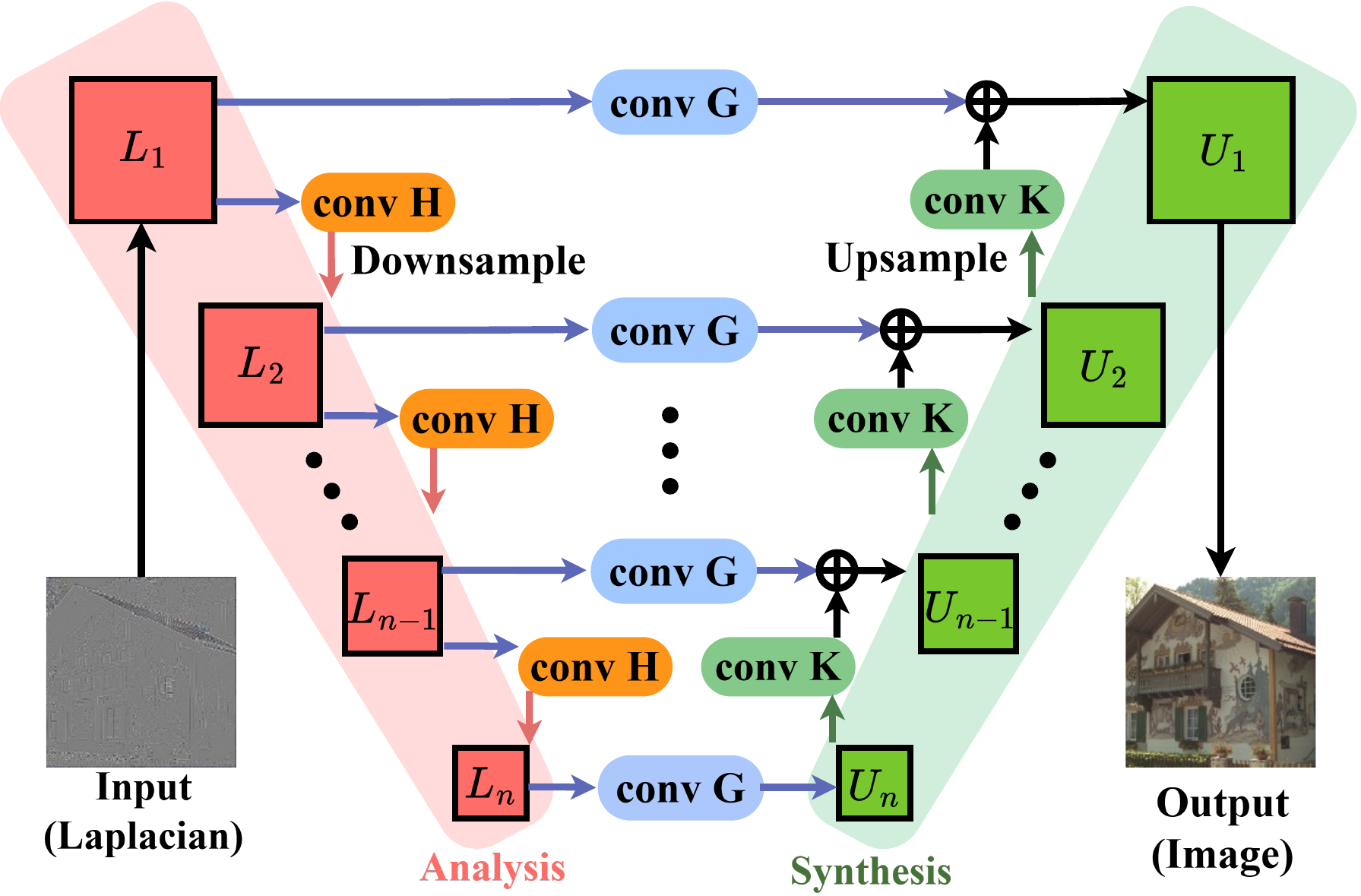}
	\caption{Shared-kernel wavelet network. The learnable kernels $H$, $G$, and $K$ are reused across all scales, while all intermediate states remain three-channel representations. The nonlearned per-channel DC correction is omitted for clarity.}
	\label{fig:shared_kernel_network}
\end{figure}

We consider the shared-kernel wavelet network in Fig.~\ref{fig:shared_kernel_network} for reconstructing an image from a Laplacian field $\mathbf{y}$ with given per-channel DC information~\cite{gong2026sharedkernel}. Let $\mathcal{P}$ denote zero padding and $\mu(\cdot)$ the per-image, per-channel spatial mean. Let $\mathbf{d}=\mu(\mathcal{P}(\mathbf{x}))\in\mathbb{R}^{3}$ denote the per-channel DC vector of the zero-padded reference image, where $\mu(\cdot)$ is the spatial mean. For uncorrected output $a_{\boldsymbol{\theta}}(\mathbf{y})$, the final reconstruction is
\begin{equation}
	\hat{\mathbf{x}}
	=
	F_{\boldsymbol{\theta}}(\mathbf{y},\mathbf{d})
	=
	\mathcal{C}\!\left[
	a_{\boldsymbol{\theta}}(\mathbf{y})
	-\mu\!\left(a_{\boldsymbol{\theta}}(\mathbf{y})\right)
	+\mathbf{d}
	\right],
	\label{eq:dc_correction}
\end{equation}
where $\mathcal{C}$ removes the padded boundary. The same three-scalar DC information is provided to BP and CMA-ES in all stages.

The multiscale analysis--synthesis network reuses three learnable kernels, $H$, $G$, and $K$, across all resolutions, making its trainable parameter space independent of the number of scales while preserving a highly compact parameterization.

The learnable kernels are implemented as bias-free channel-wise grouped convolutions, with $H\in\mathbb{R}^{3\times1\times5\times5}$, $G\in\mathbb{R}^{3\times1\times3\times3}$, and $K\in\mathbb{R}^{3\times1\times5\times5}$. Thus,
\begin{equation}
	\label{eq:hgk}
	d_c=5^2+3^2+5^2=59,\qquad
	\boldsymbol{\theta}\in\mathbb{R}^{3d_c}=\mathbb{R}^{177},
\end{equation}
where $d_c$ is the number of learnable parameters per RGB channel. Hence, $\boldsymbol{\theta}=[\boldsymbol{\theta}_R,\boldsymbol{\theta}_G,\boldsymbol{\theta}_B]$ with $\boldsymbol{\theta}_c\in\mathbb{R}^{59}$.

\subsection{Population-Based Forward-Only Search}
\label{subsec:population_forward_search}

Given training pairs $\{(\mathbf{y}_i,\mathbf{x}_i)\}_{i=1}^{N}$ with corresponding DC information $\mathbf{d}_i=\mu(\mathcal{P}(\mathbf{x}_i))$, the reconstruction objective is
\begin{equation}
	\label{eq:reconstruction_loss}
	\mathcal{L}(\boldsymbol{\theta})=
	\frac{1}{N}\sum_{i=1}^{N}
	\operatorname{MSE}\!\left(
	F_{\boldsymbol{\theta}}(\mathbf{y}_i,\mathbf{d}_i),
	\mathbf{x}_i
	\right).
\end{equation}
Conventional BP minimizes this objective using gradients with respect to $\boldsymbol{\theta}$. In contrast, we optimize complete parameter candidates using the Covariance Matrix Adaptation Evolution Strategy (CMA-ES)~\cite{hansen2016cma}. At each generation, candidates are sampled as
\begin{equation}
	\label{eq:cma_sampling}
	\boldsymbol{\theta}^{(k)}
	=
	\mathbf{m}
	+\sigma\mathbf{C}^{1/2}\mathbf{z}^{(k)},
	\qquad
	\mathbf{z}^{(k)}\sim\mathcal{N}(\mathbf{0},\mathbf{I}),
\end{equation}
where $\mathbf{m}$, $\mathbf{C}$, and $\sigma$ denote the search mean, covariance matrix, and global step size, respectively. Candidates are ranked using forward reconstruction losses, after which CMA-ES adapts its search distribution. The learning loop therefore requires no automatic differentiation or backward propagation.

\subsection{Structure-Aware Search and Credit Assignment}
\label{subsec:structure_aware_search}

Although the overall parameter space is compact, directly searching $\boldsymbol{\theta}\in\mathbb{R}^{177}$ ignores the intrinsic RGB separability of the grouped network. We therefore factorize the search into three channel-specific distributions,
\begin{equation}
	\label{eq:channel_decomposition}
	q_c(\boldsymbol{\theta}_c)
	=
	\mathcal{N}
	\left(
	\mathbf{m}_c,\sigma_c^2\mathbf{C}_c
	\right),
	\qquad
	\boldsymbol{\theta}_c\in\mathbb{R}^{59},
	\quad
	c\in\{R,G,B\}.
\end{equation}
At each population index, one candidate from each channel distribution is assembled into a network candidate and evaluated through the same RGB forward pass. Because all learnable convolutions are channel-wise grouped and the remaining operations preserve channels, each output channel depends only on its corresponding parameter block.

For consistent ranking, each coverage partitions all $N$ samples into non-overlapping subsets of size $M$. At generation $t$, all candidates and three CMA-ES instances share $\mathcal{S}_t$; the permutation is refreshed every $N/M$ generations.

We further use channel-specific credit assignment so that each parameter block receives its directly corresponding optimization signal. For channel $c$, the fitness is
\begin{equation}
	\label{eq:channel_loss}
	\mathcal{L}_c=
	\frac{1}{M}\sum_{i\in\mathcal{S}_t}
	\operatorname{MSE}\!\left(
	F_{\boldsymbol{\theta}}(\mathbf{y}_i,\mathbf{d}_i)_c,
	\mathbf{x}_{i,c}
	\right),
	\quad c\in\{R,G,B\}.
\end{equation}
The $R$, $G$, and $B$ CMA-ES instances are ranked and updated using $\mathcal{L}_R$, $\mathcal{L}_G$, and $\mathcal{L}_B$, respectively. This combination of parameter-space decomposition and aligned credit assignment allows population-based search to exploit the intrinsic RGB-separable structure while retaining a fully forward-only learning procedure.

\section{Experiments}
\label{sec:experiments}

We evaluate the framework through comparisons with backpropagation, structure and credit-assignment ablations, and computational analysis. We first describe the common protocol used across comparisons and multi-seed experiments.
\subsection{Experimental Setup}
\label{subsec:experimental_setup}
Experiments use a BSDS500-derived reconstruction dataset. We extract $256\times256$ patches from the original training, validation, and test splits and apply eight-way rotation-and-flip augmentation, yielding 51,200, 1,280, and 12,800 samples, respectively, without source-image overlap. MSE, PSNR, and SSIM are used for evaluation. All methods share the same 177-parameter network, Xavier initialization, data splits, and forward implementation; validation selects checkpoints and the test set remains isolated. The BP baseline uses Adam with a learning rate of $10^{-5}$, batch size 32, and 1,500 epochs over ten data-order seeds. CMA-ES uses $\sigma_0=0.25$, $M=64$ samples per generation, and 4,800 generations, corresponding to six training-set coverages. We use $\lambda=32,64$ for the main comparison and $\lambda\in\{19,32,48,64\}$ for the population study. The choice $M=64$ follows a preliminary ranking-stability analysis, balancing reliability and forward-evaluation cost.

\subsection{Comparison with Backpropagation}
\label{subsec:bp_comparison}

Table~\ref{tab:main_comparison} compares the proposed forward-only learner with Adam/BP. Values are means over ten runs. Without backward propagation, both forward-only configurations outperform BP in reconstruction quality. In particular, $\lambda=64$ reduces the average test MSE from $4.1783$ to $4.0130$, corresponding to a $3.95\%$ reduction, while also improving PSNR and SSIM.

\begin{table}[t]
	\centering
	\caption{Comparison with Adam/BP on the BSDS500 test set. Reconstruction metrics are means over ten runs. Forward evaluations denote training-image evaluations.}
	\label{tab:main_comparison}
	\small
	\setlength{\tabcolsep}{5.2pt}
	\renewcommand{\arraystretch}{1.0}
	\begin{tabular}{lcccc}
		\hline
		Method & MSE$\downarrow$ & PSNR$\uparrow$ & SSIM$\uparrow$ & Fwd. evals\\
		\hline
		Adam/BP & 4.1783 & 42.6211 & 0.994875 & 76.80M\\
		Ours ($\lambda=32$) & 4.0431 & 42.7583 & 0.995019 & 9.83M\\
		Ours ($\lambda=64$) & \textbf{4.0130} & \textbf{42.7870} & \textbf{0.995035} & 19.66M\\
		\hline
	\end{tabular}
\end{table}

The results are also highly consistent across random runs. Across ten runs, the test-MSE standard deviations of Adam/BP, Ours ($\lambda=32$), and Ours ($\lambda=64$) are $0.0010$, $0.0121$, and $0.0008$, respectively. The test-MSE coefficient of variation is $0.0241\%$ for BP and $0.0209\%$ for the $\lambda=64$ configuration. Moreover, all ten runs of both forward-only configurations obtain lower test MSE than all BP runs, indicating that the gain is not driven by isolated favorable runs.
\subsection{Structure and Credit Assignment}
\label{subsec:structure_credit}

\begin{figure}[t]
	\centering
	\includegraphics[width=\columnwidth]{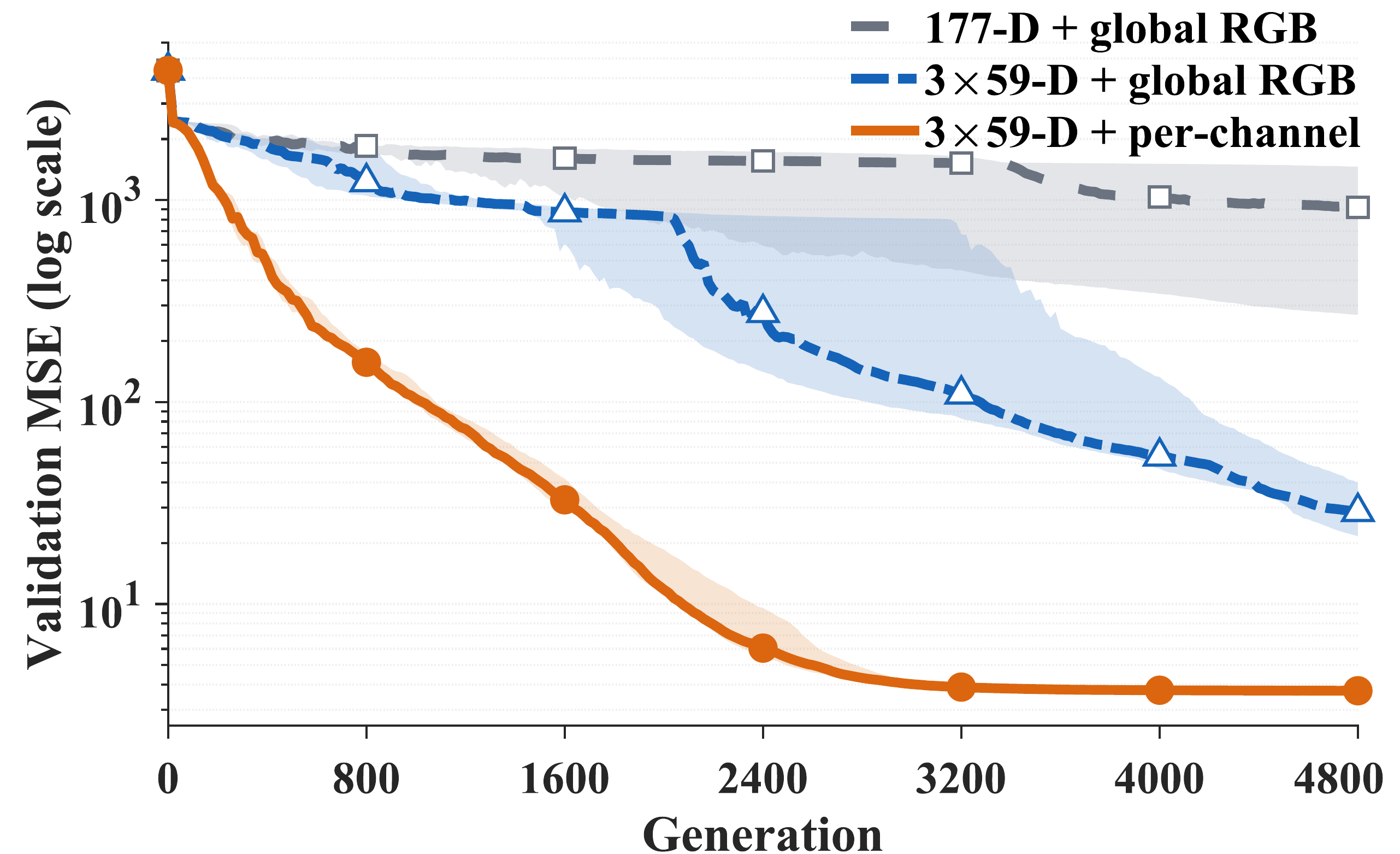}
	\caption{Mechanism ablation at $\lambda=32$. Validation-MSE trajectories compare joint 177-D search, structured $3\times59$-D search with global fitness, and channel-specific credit assignment. Curves show median and min--max envelopes over five matched seeds; all runs use $M=64$ and 4,800 generations.}
	\label{fig:mechanism_ablation}
\end{figure}

We examine how parameter-space structure and fitness assignment affect forward-only optimization. Figure~\ref{fig:mechanism_ablation} compares three formulations at $\lambda=32$: joint 177-D search with global RGB fitness, $3\times59$-D structured search with the same fitness, and $3\times59$-D search with channel-specific credit assignment. The joint search converges slowly and remains at high validation error. Decomposing the search space substantially improves convergence, while assigning each parameter block its channel-specific fitness produces a further large improvement and reaches the low-error regime.

The test MSE decreases from $940.23$ for joint 177-D search to $30.81$ after $3\times59$-D decomposition, and further to $4.04$ with channel-specific credit assignment. Thus, compactness alone is insufficient for whole-network search: parameter-space decomposition is decisive, while channel-specific credit assignment is critical for reaching BP-level reconstruction quality under the tested optimization budget. Together, the two components align the optimizer with the intrinsic RGB-separable structure of the network.

\begin{figure}[t]
	\centering
	\includegraphics[width=\columnwidth]{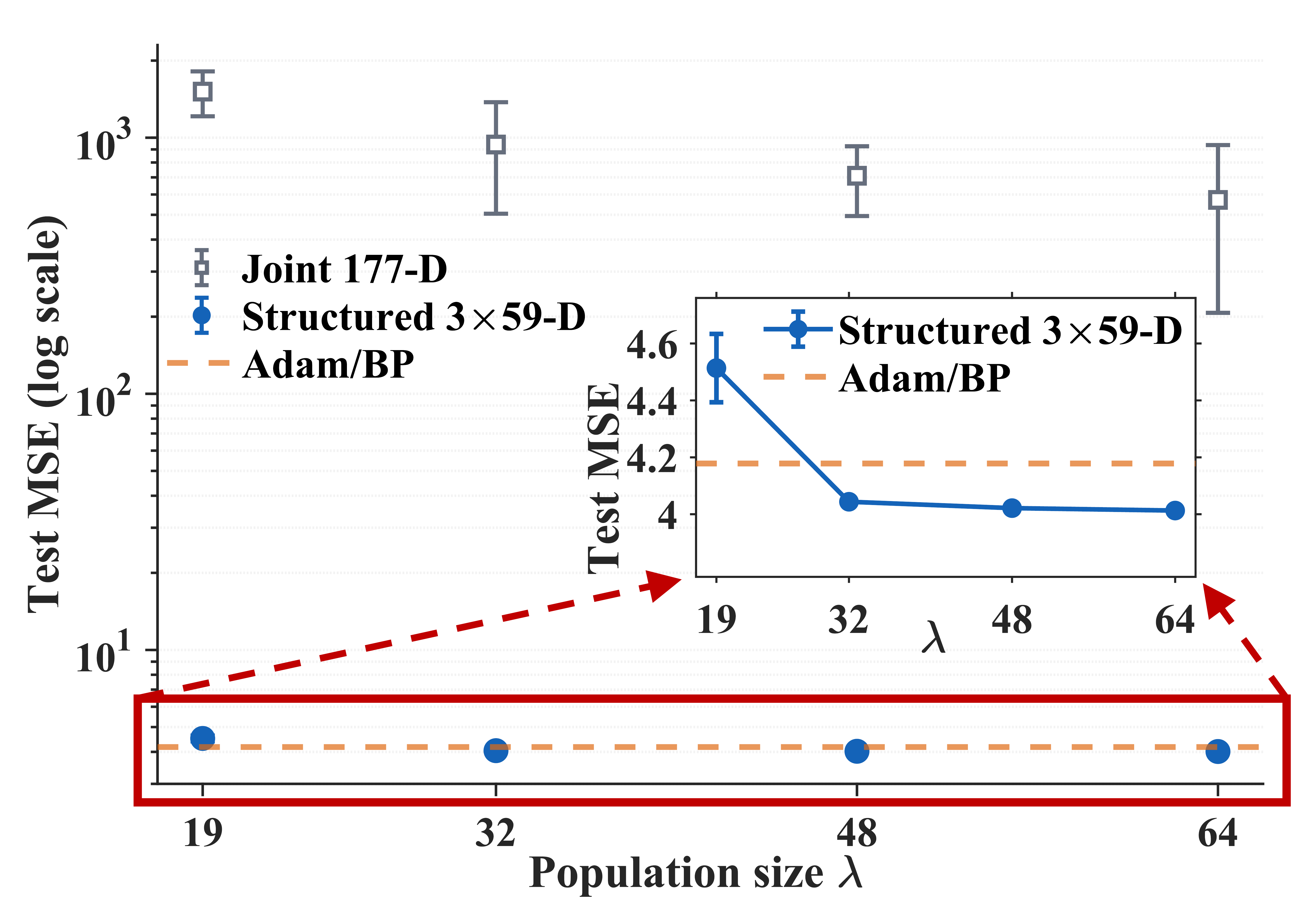}
	\caption{Population-size response of joint 177-D and structured $3\times59$-D forward-only search. The inset magnifies the structured regime with Adam/BP as a reference. Points show mean$\pm$standard deviation over five matched seeds; all runs use $M=64$ and 4,800 generations.}
	\label{fig:population_response}
\end{figure}

We next study the effect of population size. Figure~\ref{fig:population_response} compares joint 177-D and structured $3\times59$-D optimization for $\lambda\in\{19,32,48,64\}$. Increasing $\lambda$ improves both formulations, but even the joint 177-D search at $\lambda=64$ remains far from the BP regime. In contrast, the structured formulation achieves orders-of-magnitude lower error. As highlighted by the inset, the largest improvement occurs from $\lambda=19$ to $32$, after which the gain saturates. This confirms that parameter-space structure is the dominant factor in forward-only optimization, whereas increasing population provides only secondary refinement after structured search is established.

\subsection{Computational Characteristics}
\label{subsec:computational_characteristics}

As shown in Table~\ref{tab:main_comparison}, CMA-ES requires only forward evaluations, whereas BP additionally propagates gradients through all training samples. These counts are not a strict equal-compute comparison: larger $\lambda$ increases candidate evaluations, and BP and CMA-ES use different GPU parallelism. We therefore make no claim of wall-clock superiority.

\section{Conclusion}
\label{sec:conclusion}

This work presented a global forward-only learning framework for compact shared-kernel wavelet neural networks. By combining CMA-ES with structure-aware parameter decomposition and channel-specific credit assignment, the proposed method enables whole-network optimization using forward evaluations only, without automatic differentiation or backward propagation. Experiments on image reconstruction show that the resulting learner consistently outperforms the Adam/BP baseline while revealing that network structure and aligned credit assignment are more important than simply increasing the population size. These results demonstrate that extreme parameter sharing can make global forward-only learning practical for compact neural networks. Future work will investigate its extension to larger architectures and more general vision tasks.

% References should be produced using the bibtex program from suitable
% BiBTeX files (here: strings, refs, manuals). The IEEEbib.bst bibliography
% style file from IEEE produces unsorted bibliography list.
% -------------------------------------------------------------------------
\bibliographystyle{IEEEbib}
\bibliography{refs}

\end{document}